\pdfoutput=1
\documentclass[runningheads]{llncs}
\usepackage{graphicx}

\usepackage{tikz}
\usepackage{comment}
\usepackage{amsmath,amssymb} 
\usepackage{color}

\usepackage[accsupp]{axessibility}  

\begin{document}
\pagestyle{headings}
\mainmatter
\def\ECCVSubNumber{100}  

\title{Masked Autoencoders for Egocentric Video Understanding @ Ego4D Challenge 2022} 


\titlerunning{Masked Autoencoders for Egocentric Video Understanding}
%
\author{Jiachen Lei\inst{1} \and
Shuang Ma\inst{2} \and
Zhongjie Ba\inst{1} \and
Sai Vemprala\inst{2} \and
Ashish Kapoor\inst{2} \and 
Kui Ren\inst{1}}
\authorrunning{Lei et al.}
%
\institute{Zhejiang University\\
\email{\{jiachenlei, zhongjieba, kuiren\}@zju.edu.cn}\\
\and
Microsoft, Redmond, WA, USA\\
\email{\{shuama, sai.vemprala, akapoor\}@microsoft.com}\\
}
\maketitle

\begin{abstract}
In this report, we present our approach and empirical results of applying masked autoencoders in two egocentric video understanding tasks, namely, Object State Change Classification and PNR Temporal Localization, of Ego4D Challenge 2022. As team ”TheSSVL”, we ranked 2nd place in both tasks. Our code will be made available.\footnote{https://github.com/jasonrayshd/egomotion}

\end{abstract}

\section{Masked Autoencoder}
MAE\cite{he2022masked}, proposed by Kaiming, etc, has now drawn the wide interest of researchers in self-supervised learning due to its pretraining efficiency and generalization ability in various downstream tasks. Recently, two parallel works\cite{tong2022videomae}\cite{feichtenhofer2022masked} called VideoMAE are proposed to extend MAE from image to video domain. In this report, we demonstrate that merely pretraining the model on 3rd-person view datasets (e.g. Kinetics 400\cite{kay2017kinetics}) under the settings of VideoMAE can achieve state-of-the-art performance on egocentric video understanding tasks including Ego4d object state change classification and Ego4d PNR temporal localization. This demonstrates the great representation learning and generalization ability of VideoMAE in self-supervised video pretraining.

\section{Downstream-Task Formulation}

Following Ego4d official baseline setting on object state change classification (OSCC) and PNR temporal localization tasks\cite{grauman2022ego4d}, we simply formulate OSCC and temporal localization tasks as a binary and a multi-class classification problem respectively. We finetune our model on these two tasks under a multi-objective optimization setting. Therefore, our model is composed of a shared backbone and two separate linear heads each used for OSCC or temporal localization. Besides, our optimization objective can be represented by equation 1, where $\theta$ denotes model parameters. $\lambda1$ and $\lambda2$ are constants and are both set to 1 in all our experiments.

\begin{align} 
\arg \min_{\theta}\ \ L = \arg \min_{\theta}\ \ \lambda1 \times L_{oscc} + \lambda2 \times L_{tl}
\end{align}

 Instead of predicting index of the PNR frame only when a state change occurs in given input frames as in \cite{grauman2022ego4d}, We treat the temporal localization task as a $ N+1$ classification problem where N denotes the number of input frames. Therefore, the classification head for temporal localization is obliged to predict whether state change occurred in given input frames and occurred in which frame if do. We believe this helps smooth the model training process and improve model performance by providing sufficient supervised information.

\section{Experiments}


\subsection{Implementation details}
In our experiments, we use ViT as our backbone of which weights are initialized from VideoMAE\cite{tong2022videomae} pretrained on Kinetics-400. Specifically, we use ViT-L pretrained for 1600 epochs in OSCC and ViT-B pretrained for 800 epochs in temporal localization. Pretraining details and model checkpoints can be found in \cite{tong2022videomae}. We will show that even with weights obtained on 3rd-person view datasets, VideoMAE shows great generalization ability on egocentric downstream tasks and surpass most existing methods both on OSCC and temporal localization tasks.

When finetuning on OSCC and temporal localization tasks, we sample 16 frames from each clip as input and set the training epochs to 100. More finetuning details are listed in Table \ref{table:ftsetting}.

During inference, we sample 3 clips out of 16 frames from each test video via spatially random cropping. The spatial dimension of obtained clips is 224x224. The final results are the average of model raw logits on these 3 clips. We submit the test result of the model checkpoint that performs best on the validation set.

\subsection{Results}
 For a more comprehensive comparison on OSCC and temporal localization tasks, we cite results both from the leaderboard and publicly available technical reports of other teams. As shown in Table \ref{table:oscc} and Table \ref{table:tl}, by simply pretraining on Kinetics 400 under the settings of VideoMAE, we ranked 2nd place in both tasks. 
 
 It is worth noting that in PNR temporal localization task, always selecting the frame corresponding to 0.45 times the video duration achieves an absolute temporal error of 0.67, as stated in \cite{escobar2022video}. This indicates methods including ours struggle to precisely predict the PNR frame.

\begin{table}
    \caption{Results on Ego4d challenge 2022 Object State Change classification task. Each method marked by * is the latest submissions of the team and the results are different from the one in technical report }
    \begin{center}
        \setlength{\tabcolsep}{10pt}
        \renewcommand{\arraystretch}{1.1}
        \begin{tabular}{lccc}
         \multicolumn{1}{l}{\bf Method} & \multicolumn{1}{c}{\bf Extra data} & \multicolumn{1}{c}{\bf Accuracy $\uparrow$} \\
         \hline
        Always positive   &       -         &   47.7                    \\
        \hline
        Uniandes-Google\cite{escobar2022video}  &       K600        &   68    \\
        TarHeels\cite{islam2022object}          &     IN-21K        &   72    \\
        Egocentric VLP\cite{lin2022egocentric}  &  IN-21K+BC+EClip  &   73.7  \\
        \hline
        SViT*\cite{ben2022structured}           &       N/A         &   75    \\
        TarHeels*\cite{islam2022object}         &       N/A         &   76    \\
        PearRepublic                            &       N/A         &   76    \\
        ohohoh                                  &       N/A         &   80    \\
        \hline
        {\bf Ours}        & K400    & 77.7 \\

        \end{tabular}
    \end{center}
\label{table:oscc}
\end{table}

However, team "ohohoh" achieves a better result and surpasses other methods by a large margin. In addition to model performance, We suppose that this gap might also root in the frame sampling strategy. Specifically, when adopting random sampling strategy following official codes, the PNR frame in each clip is not guaranteed to be sampled both in training and inference time. Therefore, it brings noise into the model training process and introduces an absolute time shift away from the ground-truth PNR frame in both training and inference time.

\begin{table}
    \caption{Results on Ego4d challenge 2022 PNR temporal localization task. Two simple baseline methods are presented, including always predicting center frame as the PNR frame and always predicting frame at 0.45 times total video duration as the PNR frame}
    \begin{center}
        \setlength{\tabcolsep}{5pt}
        \renewcommand{\arraystretch}{1.1}
        \begin{tabular}{lccc}
         \multicolumn{1}{l}{\bf Method} & \multicolumn{1}{c}{\bf Extra data} & \multicolumn{1}{c}{\bf Absolute temporal error$\downarrow$} \\
         \hline
        Always Center Frame     &       -         & 1.06   \\
        Always at 0.45 duration &       -         & 0.67   \\
        \hline
        Uniandes-Google\cite{escobar2022video}    & K600                  & 0.66   \\
        icego                                     & N/A                   & 0.66   \\
        SViT\cite{ben2022structured}              & K400, Ego4D, 100DOH   & 0.660  \\
        Egocentric VLP\cite{lin2022egocentric}    & IN-21K+BC+EClip       & 0.666  \\
        ohohoh                                    & N/A                   & 0.56   \\
        \hline
        {\bf Ours}          & K400    & 0.654  \\

        \end{tabular}
    \end{center}
    \label{table:tl}
\end{table}

During training, at first, each 30FPS 8-second clip is uniformly trimmed to between 5 and 8 seconds and the required number of frames (e.g. 16 frames) will then be sampled from it. If we denote the number of frames in a trimmed clip as $S$, its probability density function can be written as:  

\begin{align} 
    p(S)=\left\{
        \begin{array}{ll}
          \dfrac{1}{90}\ \ if\ 150<=S<=240\\
          \\
          \ 0 \ \ \ else
        \end{array}
      \right.
\end{align}

Among all the sampled frames, the frame that is closest to the ground-truth PNR frame is taken as a pseudo-PNR frame and used as a hard label when computing per-frame classification loss. if we sample 16 frames from each trimmed clip as model input, then, the expectation of absolute time shift away from the ground-truth PNR frame is $E(\frac{S}{16*30*2}) ~= 0.2\ s $. This non-negligible shift has a negative impact on model performance and will lead to a loss of precision. The same is true during inference time.

\begin{table}
    \caption{Finetuning setting. We follow the linear scaling rule: $lr=base\_lr\times batchsize/256$ while warmup learning rate is not affected.}
    \begin{center}
        \setlength{\tabcolsep}{10pt}
        \renewcommand{\arraystretch}{1.1}
        \begin{tabular}{ll}
         \multicolumn{1}{l}{\bf Hyperparameters} & \multicolumn{1}{l}{\bf Value}\\
         \hline
        Optimizer            & AdamW               \\
        Base learning rate   & 5e-4                \\
        Weight decay         & 0.05                \\ 
        Adam $\beta$         & (0.9, 0.999)        \\
        Batch size           & 8 or 32             \\
        Learning rate sched  & Cosine decay        \\
        Training epochs      & 100                 \\
        Warmup learning rate & 1e-6                \\
        Warmup epochs        & 5                   \\
        Layer-wise lr decay  & 0.75                \\
        \hline
        Patch size           & 16x16               \\
        Tubelet size         & 2                   \\
        Loss weight          & (1, 1)              \\
        \hline
        Input size           & 224x224             \\
        Augmentation         & RandAugment(9, 0.5) \\
        Drop path            & 0.1                 \\
        Mixup                & 0.8                 \\
        Cutmix               & 1.0                 \\
        Label smoothing      & 0.1                 \\

        \end{tabular}
    \end{center}
    \label{table:ftsetting}
\end{table}

\section{Conclusions}
Although we did not achieve the best performance on OSCC and temporal localization tasks in Ego4d Challenge 2022, we believe that, by paying much more attention to downstream task formulation and optimization, models that are pretrained on egocentric datasets under the settings of VideoMAE will further improve state-of-the-art performance on various Ego4d  tasks. In the meantime, we will go a step further in egocentric video learning based on masked autoencoding and are glad to share our work later.

%
%
\bibliographystyle{splncs04}
\bibliography{eccv2022submission}
\end{document}